\documentclass[letterpaper, 10 pt, conference]{ieeeconf}
\IEEEoverridecommandlockouts
\usepackage{amsmath,amssymb,amsfonts}
\usepackage{algorithmic}
\usepackage{graphicx}
\usepackage{textcomp}
\usepackage{xcolor}
\usepackage{algorithm}
\usepackage{array}
\usepackage[caption=false,font=normalsize,labelfont=sf,textfont=sf]{subfig}
\usepackage{stfloats}
\usepackage{url}
\usepackage{verbatim}
\usepackage{times}
\usepackage{epsfig}
\usepackage{float}
\usepackage{wrapfig}
\usepackage{bm,xspace}
\usepackage{comment}
\usepackage{multirow}
\usepackage{balance}
\usepackage{booktabs}
\usepackage{etoolbox,siunitx}
\usepackage{calc}
\usepackage{pifont,hologo}
\usepackage{nicefrac}
\usepackage{makecell}
\usepackage{svg}
\usepackage{enumerate}
\makeatletter
\let\NAT@parse\undefined
\makeatother
\usepackage[colorlinks=black,citecolor=green]{hyperref}
\usepackage{adjustbox}

\def\BibTeX{{\rm B\kern-.05em{\sc i\kern-.025em b}\kern-.08em
    T\kern-.1667em\lower.7ex\hbox{E}\kern-.125emX}}

\title{\LARGE \bf Blending Distributed NeRFs with Tri-stage Robust Pose Optimization}

\author{Baijun Ye$^{*1,2}$, Caiyun Liu$^{*1}$, Xiaoyu Ye$^{1,2}$, Yuantao Chen$^{1,3}$, Yuhai Wang$^{4}$,  \\Zike Yan$^{1}$, Yongliang Shi$^{1\dag}$, Hao Zhao$^{1}$, Guyue Zhou$^{1}$
\thanks{* Equal contribution.  $^{1}$Institute for AI Industry Research (AIR), Tsinghua University,$^{2}$Beijing Institute of Technology, $^{3}$Xi'an University of Architecture and Technology, $^{4}$University of Southern California.}%
\thanks{$\dag$ Corresponding author. shiyongliang@air.tsinghua.edu.cn}%
\thanks{Sponsored by Tsinghua-Toyota Joint Research Fund (20223930097).}
}

\begin{document}
\maketitle
\thispagestyle{empty}
\pagestyle{empty}

\begin{abstract}

Due to the limited model capacity, leveraging distributed Neural Radiance Fields (NeRFs) for modeling extensive urban environments has become a necessity. However, current distributed NeRF registration approaches encounter aliasing artifacts, arising from discrepancies in rendering resolutions and suboptimal pose precision. These factors collectively deteriorate the fidelity of pose estimation within NeRF frameworks, resulting in occlusion artifacts during the NeRF blending stage.
In this paper, we present a distributed NeRF system with tri-stage pose optimization. In the first stage, precise poses of images are achieved by bundle adjusting Mip-NeRF 360 with a coarse-to-fine strategy. In the second stage, we incorporate the inverting Mip-NeRF 360, coupled with the truncated dynamic low-pass filter, to enable the achievement of robust and precise poses, termed Frame2Model optimization. On top of this, we obtain a coarse transformation between NeRFs in different coordinate systems. In the third stage, we fine-tune the transformation between NeRFs by Model2Model pose optimization. After obtaining precise transformation parameters, we proceed to implement NeRF blending, showcasing superior performance metrics in both real-world and simulation scenarios. Codes and data will be publicly available at \href{https://github.com/boilcy/Distributed-NeRF}{https://github.com/boilcy/Distributed-NeRF}.
\end{abstract}

\section{INTRODUCTION}
The field of large-scale scene modeling has garnered considerable scholarly interest, notably in light of the recent advent of NeRF. This emergence is primarily attributed to NeRF's capacity for achieving photorealistic rendering while maintaining a compact model structure. This has brought NeRF to the forefront of attention in drone mapping~\cite{meganerf} and navigation~\cite{nerfnav}.

In the domain of large-scale scene representation using NeRF, there exist primarily two predominant methodologies. 1) \textbf{Batch learning}, as exemplified by Bungee-NeRF~\cite{bungeenerf}, requires substantial computational resources, especially for the exhaustive sampling and fitting of the entire scene. To mitigate the substantial computational requirements associated with batch training on large-scale data, there exist methods~\cite{blocknerf}~\cite{meganerf} of subdividing a vast scene into several smaller scenes. The prerequisite for this approach is that all images to be trained must share a common coordinate system. In GNSS denied environment, we often rely on methods such as Simultaneous localization and mapping (SLAM)~\cite{campos2021orb} or structure-from-motion (S$f$M)~\cite{lindenberger2021pixel}~\cite{colmap} for pose estimation. For large-scale urban scenes, SLAM inevitably accumulates errors over time, making it unreliable for precise pose information. Meanwhile, S$f$M undoubtedly demands substantial computational resources. 
2) \textbf{Incremental learning} may lead to insufficient scene representation due to forgetting. Besides, current approaches using explicit encoding methods like grid~\cite{nice}~\cite{rosinol2022nerf}  and octree\cite{vox-fusion} for real-time performance, which face the challenge of exponentially expanding encoding components as the scene scale increases, leading to substantially increased storage requirements.


We aim to devise a large-scale NeRF system under computational constraints. Our batch learning approach focuses on mitigating the forgetting problem inherent to incremental learning-based NeRF in expansive urban settings. Additionally, it efficiently handles the computational complexities arising from the increased data volumes in these scenarios. However, we encounter a \textbf{challenge} in achieving precise registration results for distinct NeRFs acquired by different agents using heterogeneous coordinate systems.

To address these limitations, we put forward a distributed NeRF framework with a tri-stage pose optimization methodology. We use Mip-NeRF 360~\cite{mipnerf360} as backbone due to its good anti-aliasing effect. In the first stage, drawing inspiration from BARF~\cite{barf}, we enhance Mip-NeRF 360 to implement the bundle-adjusting Mip-NeRF 360. A coarse-to-fine approach is employed for the conjoint optimization of scene representation and pose. In the second stage, taking cues from LATITUDE~\cite{latitude}, we leverage the principles of Truncated Dynamic Low-pass Filter (TDLF) to refine the inverting Mip-NeRF 360, termed iMNeRF. This method is similar to blurring images to make the optimization process more robust, thereby enabling pose optimization for Frame2Model. Subsequently, we employ a co-view region retrieval method (detailed in section \ref{co-view}) to search the most analogous images across diverse NeRF instances, subsequently identifying their associated poses.
Given the associated poses, we employ the iMNeRF to optimize these poses by photometric losses among rendering images and observed images, thereby obtaining reliable Frame2Model transformations. In the third stage, we obtain a rough Model2Model transformation between NeRFs by different Frame2Model transformations. Then, we project the different NeRF models onto a unified coordinate system and further optimize the relative transformations among NeRFs using the rendered images as observation, that is through the Model2Model optimization to obtain the precise transformations among NeRFs. Utilizing the tri-stage pose optimization, we implement the NeRF blending and get better performance. To validate our method, we have concurrently released both real-world and simulation datasets, demonstrating the superiority of our approach.
In summary, our contributions are as follows:
\begin{itemize}
\item[$\bullet$] We introduce a distributed NeRF framework for large-scale urban environments, incorporating a tri-stage pose optimization. This is specifically utilized during NeRF blending to address the issue of misalignment caused by inaccuracies in registration.
\item[$\bullet$] We implement bundle-adjusting Mip-NeRF 360, facilitating a conjoint optimization of poses and scene representation. Building upon this, an enhanced Frame2Model pose estimation technique, iMNeRF, is proposed. This not only optimizes registration results but also provides a dependable preliminary Model2Model transformation.
\item[$\bullet$] We release a comprehensive dataset that combines amalgamates both real-world and simulation data. The superior blending and registration results of our methodology are distinctly showcased.
\end{itemize}

\section{Related Work}
\subsection{NeRF2NeRF Registration}
NeRFs are optimized from accurately posed images. In city reconstruction with UAVs, GPS or RTK-equipped UAVs are commonly employed. However, the initial poses obtained from GPS/RTK often require refinement using S$f$M~\cite{colmap} techniques for each sub-area to achieve high-quality large-scale reconstructions. This refinement process will result in a coordinate system with arbitrary global positions that are specific to each NeRF submodule.

While extensive research exists on registration methods for explicit representations such as point clouds, there is a notable lack of studies addressing NeRF2NeRF registration for implicit fields. Recent approaches like nerf2nerf~\cite{goli2022nerf2nerf} and Zero-NeRF~\cite{peat2022zero} rely on extracting surfaces from NeRF representations for pairwise registration. NeRFuser~\cite{fang2023nerfuser} applies an off-the-shelf S$f$M method on re-rendered images to get transformation between different NeRFs. However, these methods primarily rely on traditional geometry-based methods. The storage of explicit geometric information grows infinitely as the scene expands, while implicit maps can directly store the color and geometric information of the scene through compact neural network representations, achieving registration more concisely. Further, their applicability is mainly limited to object-level or small-scene NeRFs and may not be suitable for large-scale city scene reconstructions. 
Our method utilizes NeRF to directly optimize transformations between NeRFs, achieving state-of-the-art (SOTA) performance on large-scale city scenes.
\subsection{NeRF for Large-scale Scene}
The neural radiance field has shown immense potential in representing large-scale scenes. Some methods have been proposed to leverage this technology in city scene reconstruction. 
Block-NeRF~\cite{blocknerf} spatially decomposes the scene into independently trained NeRFs. It employs Inverse Distance Weighting (IDW) and visibility prediction to calculate their contributions to the overall scene representation.
Mega-NeRF~\cite{meganerf} goes one step further by applying a geometric clustering algorithm to partition training pixels into different NeRF submodules.
Switch-NeRF~\cite{mi2023switchnerf} leverages a Sparsely Gated Mixture of Experts (MoE) approach for end-to-end learning-based scene decomposition.
However, these approaches require known poses in each scene partition under a common global coordinate system, which is impractical for large-scale scenes where acquiring accurate poses for the entire area at once is unfeasible.
Grid-NeRF~\cite{xu2023gridguided} and GP-nerf~\cite{zhang2023efficient} use efficient feature grids for scalability but still face network capacity limitations for extremely large scenes. Our method introduces the possibility of high-quality NeRF reconstruction for infinitely large-scale scenes.

\section{Formulation}
Our goal is to accurately register distributed NeRF models to a global coordinate and make full use of all NeRF models to render images. Pose optimization is divided into three stages: First, local poses are refined while training, e.g. agent $i$ collects image set $\{I_i^{C_k}\}$ for NeRF model $\mathcal{F}_i$, where $C_k$ presents a camera model and $k$ is the camera index in local dataset. Each image has a corresponding imperfect camera pose $\{T_i^{C_k}\}$ obtained from COLMAP~\cite{colmap}. To match the scene images and poses more accurately, bundle adjusting is needed, which can be expressed as eq. (\ref{eq:f1}),
\begin{equation}
\min_{T_i^{C_1},T_{i}^{C_2},\dots,T_i^{C_m}, \mathbf{\Theta}}\sum_{k=1}^{m}\left\|\hat{I}(\mathcal{F}_i(T^{C_k}_i;\boldsymbol{\Theta}))-I_{i}(\mathbf{x})\right\|_{2}^{2}.
\label{eq:f1}
\end{equation}

In the second stage, Frame2Model pose optimization is implemented through a pose gradient updating method. Since different agents define their own local coordinates, a transformation matrix from local to global is required to achieve seamless blending between models.
Translation matrix $T_{ij}$ between $\mathcal{F}_i$ and $\mathcal{F}_j$ should satisfy eq. (\ref{eq:transformmatrix}),
\begin{align}
\label{eq:transformmatrix}
    T_{ij}^* =& \min_{T_{ij}} \sum_{k=1}^{n_i} \left\|\hat{I}\left(\mathcal{F}_j(T_{ij}^{-1} T_i^{C_k} )\right)-I_i^{C_k}\right\|_2^2 \\
    &+ \sum_{k=1}^{n_j} \left\|\hat{I}\left(\mathcal{F}_i(T_{ij} T_j^{C_k} )\right)-I_j^{C_k}\right\|_2^2 \notag
\end{align}

In the third stage, Model2Model pose optimization is performed before NeRF blending, with query camera pose in a specified coordinate, e.g. $T_i$ in coordinate of agent $i$, we again optimize $T_{ij}$ with previous $T_{ij}^*$ as a initial value:
\begin{align}
\label{eq:beforeblending}
    T_{ij}^* =& \min_{T_{ij}} \left\|\hat{I}\left(\mathcal{F}_j((T_{ij}^{-1} T_i)\right)-\hat{I}\left(\mathcal{F}_i(T_i)\right)\right\|_2^2.
\end{align}

\section{Method}

\begin{figure*}[!t]
\centering
\includegraphics[width=1\textwidth]{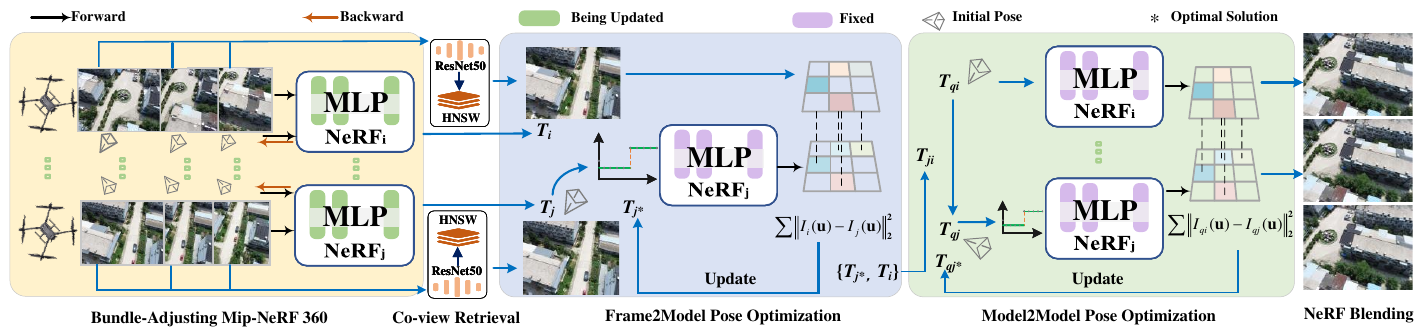}%
\caption{System Overview. Each agent trains a bundle-adjusting Mip-NeRF 360. Following this, co-view region retrieval provides an initial value for Frame2Model pose optimization. A final Model2Model pose optimization is performed before blending to achieve a seamless fusion of NeRFs.}
\label{main}
\vspace{-3mm}
\end{figure*}

\subsection{System Overview}
As shown in Fig.\ref{main}, the system includes two tasks: NeRF registration and NeRF blending. In the NeRF registration stage, rather than directly addressing equation \ref{eq:transformmatrix}, we adopt a tri-stage pose optimization. Initially, bundle-adjusting Mip-NeRF 360 is introduced to achieve joint optimization of both scene representation and poses of input images. Subsequently, employing the co-view region retrieval approach, we identify similar image pairs like $(I_i^{C_k}, I_j^{C_t})$ in each agent's dataset, along with their local coordinate poses $(T_i^{C_k}, T_j^{C_t})$. Beginning with $T_j^{C_t}$ as the initial guess of $T_j^{C_k}$, the pre-trained NeRF model minimizes the disparity between the rendered image $I_j^{C_k}$ and the $I_i^{C_k}$. 
A coarse-to-fine strategy ensures avoidance of local optima when performing Frame2Model optimization, and Frame2Model registration results are obtained. 
Thirdly, coarse relative transformations between NeRFs obtained from Frame2Model registration results are regarded as prior to transform sampled points of different NeRFs to the unified coordinate system. 
After determining $T_j^{C_k}$, we ascertain the transformation matrix $T_{ij}$. 
Initiated with a query camera pose $T_i^{C_q}$, optimization before rendering entails iterating the previous pose optimization with fewer steps and excluding TDLF. After registration, the contribution of each NeRF can be determined using inverse distance weighting (IDW). In the far right of the figure, from top to bottom, the first two images are rendered by two distributed NeRFs from the same pose. The bottom most image is the result of blending.

\subsection{Bundle-Adjusting NeRF for Pose Refinement}
\label{NeRF_doc}
As a compact representation of scenes, Mip-NeRF 360 adeptly addresses aliasing issues during rendering at varying resolutions and the ambiguity associated with reconstructing 3D content from 2D images in large-scale unbounded environments.
However, even though the camera extrinsic obtained from S$f$M or RTK are precise, errors are inevitable, which consequently affects the quality of the scene representation. Therefore, given the imperfect camera extrinsic and the corresponding images, our method is to simultaneously optimize the scene representation and camera extrinsic by minimizing a weighted combination of warp loss~\cite{barf}, distortion loss~\cite{mipnerf360} and proposal loss~\cite{mipnerf360}:
\begin{equation}
   \label{nerf}
     {\Theta ^*},{T ^*} =\mathcal{L}_{warp}(\cdot)+\lambda\mathcal{L}_{dist}(\cdot)
     +\mathcal{L}_{prop}(\cdot),
\end{equation}
which are respectively responsible for the alignment of pose, scene geometry and training efficiency.

During training, applied coarse-to-fine optimization mainly emphasise positional encoding. Mip-NeRF 360 constructs an integrated positional encoding (IPE) representation of the volume covered by each conical frustum cast from rendering pixel, instead of constructing positional encoding (PE) features for sampled points along the casted line.
While this approach yields effective anti-aliasing performance, pose optimization leveraging this model may also be trapped in a local optimum due to the disproportionate impact of high-frequency components on the gradient.
Drawing inspiration from BARF~\cite{barf}, we introduce a coarse-to-fine optimization strategy by adding a weight layer $\omega(\alpha)$, where the weight $\omega_k$ of $k$-th frequency component is:
\begin{equation}
    \omega_k(\alpha)= \begin{cases}0 & \text { if } \alpha<k \\ \frac{1-\cos ((\alpha-k) \pi)}{2} & \text { if } 0 \leq \alpha-k<1 \\ 1 & \text { if } \alpha-k \geq 1\end{cases}
\end{equation}
Let $\alpha \in [0, L]$ be modulated in relation to the training progression, as illustrated in Fig.\ref{low-pass-filter}(a). When $\omega_k(\alpha)$ is set to 0, contributions to the gradient from the $k$-th frequency component and above are effectively neutralized. Starting with $\alpha=0$, encodings are progressively activated until full positional encoding is achieved at $\alpha=L$. This approach facilitates our model's initial alignment with a smoother signal, paving the way for the subsequent capture of a high-resolution scene representation.

\subsection{Co-view Region Retrieval}
\label{co-view}
Assuming the data collection is guaranteed to have overlapping areas, we present a co-view region retrieval approach that leverages the capabilities of deep learning for feature extraction and efficient similarity search methods. We employ the pre-trained ResNet-50 model, leveraging its established proficiency in extracting intricate feature hierarchies from images~\cite{he2016deep}. By adapting the ResNet-50 architecture and omitting its final fully connected layer, we seek a high-dimensional representation of images, summarizing their fundamental visual attributes concisely~\cite{imagenet}.
Consequently, we extract feature vectors from all images of datasets for all distributed NeRFs, and employ these vectors to establish an index within the FAISS library~\cite{faiss}. 
The Hierarchical Navigable Small World (HNSW)~\cite{hnsw}~\cite{nsw} search algorithm is adopted for its for its balanced performance in speed, precision, memory efficiency, and scalability, especially pertinent to the retrieval of high-dimensional vectors. When it comes to similarity searches, the extracted feature vector of a given query image is utilized to probe the FAISS index, identifying the pairs of images with the highest similarity from the dataset based on the L2 distance between feature vectors. 

\subsection{Frame2Model Pose Optimization}
After finding the co-view region across multiple agents, the subsequent step is computing relative transformations between distributed NeRFs. Given a specific set of image pairs $(I_i^{C_k}, I_j^{C_t})$, we extract the associated poses $T_i^{C_k}$ and $T_j^{C_t}$ which are previously optimized during the training stage of bundle-adjusting Mip-NeRF 360. Taking camera $C_k$ of $(I_i^{C_k}, T_i^{C_k})$ for example, we use iMNeRF to locate this camera in NeRF $\mathcal{F}_j$. $I_i^{C_k}$ serves as an observation and $T_j^{C_t}$ serves as an initial guess of $T_j^{C_k}$.

iMNeRF aims to optimize $T_j$ by minimizing the photometric loss between the observed image $I_i$ and the image $\hat{I}_j^{C_k}$ rendered from our NeRF $\mathcal{F}_j$. To ensure smoother optimization and faster convergence, we perform our optimization on the tangent plane. This can be formulated as follows:

\begin{equation}
    \xi^* = {\min _{\xi \in \mathfrak{se}(3)} \left\|\hat{I} \left(\mathcal{F}_j(\xi; T_j)\right)-I \right\|_2^2} 
\end{equation}
\begin{equation}
    {T_j^{C_k}} =\exp(\xi^*) {T}_j\label{Opose}
\end{equation}

\begin{figure}[!t]
\centering
\includegraphics[width=0.5\textwidth]{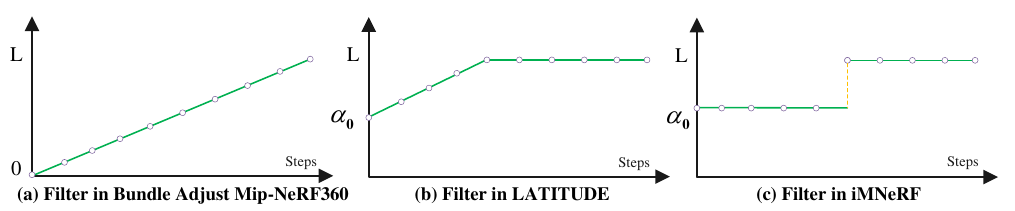}%
\vspace{-1mm}
\caption{Different Dynamic Low-pass Filter.}
\label{low-pass-filter}
\vspace{-3mm}
\end{figure}

Due to our NeRF's outstanding ability to render without image aliasing across various resolutions, our approach remains competent in producing high-quality images, even amidst altitude changes from a drone's top-down perspective. This broadens the potential applications of our model. IPE increases the accuracy of high-frequency position encoding without eliminating the scene's high-frequency information, leading to two pivotal design decisions in our method. First, the coarse-to-fine strategy, supported by \cite{barf} and \cite{latitude}, helps avoid local optima. Drawing from LATITUDE~\cite{latitude}, we incorporated a TDLF to IPE to suppress invalid outputs at high frequency, such as artifacts. Second, there's no need to sequentially modify high-frequency encoding. Instead of the filter in LATITUDE (Fig.\ref{low-pass-filter}(b)), we only need to eliminate some high frequencies at the beginning and release all high-frequency information once reaching an iteration threshold.
 The weight $\omega_k$ is:
\begin{equation}
    \omega_k(\alpha)= \begin{cases} 0 & \text { if } \alpha<k \\
    1 & \text { if } \alpha \geq k
    \end{cases}
\end{equation}
where $\alpha \in \{\alpha_0, L\}$ changes if progress exceeds a threshold $\tau = 0.8$ and the empirical value for $\alpha_0$ is 0.6 (Fig.\ref{low-pass-filter}(c)).

\subsection{Model2Model Registration}

Once the $T_j^{C_k}$ is derived from Frame2Model process, the relative transformation can be formulated as $T_{ji} = T_j^{C_k}(T_i)^{-1}$. A parallel process is employed to locate camera $C_t$ of $(I_j, T_j)$ within NeRF $\mathcal{F}_i$, yielding the transformation $T_{ij} = T_i^{C_t}(T_j)^{-1}$. 

This method is reiterated across several image pairs displaying relatively high similarity. For each iteration, we compute $\lvert T_{ji}T_{ij} - E \rvert$, where $E$ is the identity matrix. The transformation $T_{ji}$ that offers the minimum value for this metric is selected as the preliminary estimation of relative poses between the distinct NeRFs. 

However, for achieving high-fidelity rendering, a more refined transformation matrix is necessary when considering the query camera $C_q$, as noise in the observational data including inaccurate pose or blurry captures can compromise the accuracy of relative pose after the Frame2Model pose optimization. 

Therefore, we conduct a final optimization before blending. Given the pre-trained model of NeRF $\mathcal{F}_i$, we utilize it to generate a set of synthesized images $\{I_i^{C_q}\}$ by querying it with sampled camera pose $\{T_{i}^{C_q}\}$, while $\{I_j^{C_q}\}$ is rendered by pose $\{T_{j}^{C_q}\}$ that is computed by $\{T_{i}^{C_q}\}$ and previous $T_{ji}$. Again we perform iMNeRF to solve $T_{ji}^{*}$.
Upon obtaining the ultimate $T_{ji}^{*}$, the rendered images align closely, serving as a cornerstone for facilitating the blending of large-scale scenes. 

\subsection{NeRF Blending}
We use the modified inverse distance weighting (IDW) method based on \cite{fang2023nerfuser} to carry out NeRF blending. A given query camera will generate samples in multiple models and IDW considers combining the 3D points sampled by rays casted in different coordinate systems, rendering in the same frame in the context of accurate registration result. Although Mip-NeRF 360 performs sampling based on Gaussian, we can interpret the sampling results as the midpoints of intervals and subsequently aggregate them along the same ray.
After merged samples $\left\{\left(\bar{t}_k, \bar{\delta}_k\right)\right\}_k$ ($k$ is the point index on merged ray) are obtained from outputs of NeRF $\mathcal{F}_i$ and $\mathcal{F}_j$, distance from sample points to each NeRF origin are known. We selectively adjust contribution by weight $w_{k}$ from NeRF points-wise based on the distance ratio.
\begin{equation}
    w_{i,k} = \begin{cases}
    1 & \text { if } \frac{d_{i,k}}{d_{j,k}} < 0.5 \\ 
    \frac{d_{j,k}^5}{d_{i,k}^5+d_{j,k}^5} & \text { if } 0.5 \leq \frac{d_{i,k}}{d_{j,k}} < 2 \\
    0 & \text { if } \frac{d_{i,k}}{d_{j,k}} \geq 2 
    \end{cases}
\end{equation}
$w_{j,k}$ is determined in the same manner.

\section{EXPERIMENTS AND ANALYSIS}
\subsection{Implementation Details}
We implement bundle-adjusting Mip-NeRF 360, iMNeRF, and NeRF Blending based on JAX. The real scene is around 120 $\times$ 60 $m^2$. We train the scene for 250,000 iterations. The MLPs configuration of Mip-NeRF 360 configuration is slightly changed, including a proposal MLP with 4 hidden layers and 256 units, a MLP with 8 layers and 1024 hidden units. We resize the images to 1216 × 912 pixels and randomly cast rays during the training steps. Adam optimizer is adopted with an initial learning rate of $2 \times 10^{-3}$ decaying exponentially to $2 \times 10^{-5}$ for scene reconstruction and $2 \times 10^{-2}$ to $6 \times 10^{-3}$ for pose optimize. In order to implement a coarse-to-fine strategy, we add a deterministic layer after IPE, which applies the weights on the encoded feature. Experiments are conducted on a single NVIDIA RTX3090 GPU with 24GB of memory.

\subsection{Dataset}
We evaluate our method using our Distributed Urban Minimum Altitude Dataset (DUMAD), comprising both real-world and virtual scenes. The virtual scene is generated using the AirSim~\cite{airsim2017fsr} simulation built on Unreal Engine. To simulate distributed scenarios, we deploy three drones in both scenes, each with distinct starting origins resulting in disparate coordinate systems. Further, To enhance data realism, we incorporate two authentic city scene models within Unreal Engine, faithfully replicating urban environments resembling New York and San Francisco. Owing to space constraints in this paper, the visualization of the simulation experiments will be viewed in the attached supplementary video.

The real-world data is collected using a DJI M300RTK drone with traditional oblique photography settings at different times. We employ COLMAP~\cite{colmap} for pose refinement, dividing the entire scene into four areas. As far as we know, we are the first to release a dataset tailored for training distributed NeRF models in large-scale city scenes.

\subsection{Bundle-Adjusting for NeRF Representation}
In the real world, there are inherent errors in pose estimation, which are not present in simulation environments. Therefore, comparing our bundle-adjusting Mip-NeRF 360 with the original Mip-NeRF 360 in a simulation setting is not meaningful. To provide a meaningful comparison, we benchmark against NGP-based~\cite{ngp} NeRFacto~\cite{nerfstudio} (NeRFacto also operates in a joint  optimization mode of scene representation and pose), which serves as the backbone of NeRFuser and currently the most widely used NeRF method in real-world scenarios. As shown in Table \ref{tab:joint}, our method outperforms in terms of quantitative results. This ensures that we can provide more accurate and reliable prior information for the subsequent registration stage.
\begin{table}[H]
  \centering
  \caption{Performance of NeRF.}
  \small
  \renewcommand{\tabcolsep}{5pt}
 \resizebox{0.3\textwidth}{!}{
 \begin{tabular}{@{}clcccc@{}}
  \toprule[1pt]
   {\begin{tabular}[c]{@{}c@{}}Scene\end{tabular}}
   &{\begin{tabular}[c]{@{}c@{}}Methods\end{tabular}}
   &{PSNR$\uparrow$}
   &{SSIM$\uparrow$} 
   &{LPIPS$\downarrow$}  \\
\midrule
\multirow{3}{*}{\begin{tabular}[c]{@{}c@{}}Real\end{tabular}} 
    &NeRFacto    & 21.95 & 0.699 & 0.302 & \\
    &Mip-NeRF 360  &  26.25 &  0.801 & 0.224 &\\
    &Ours      & \textbf{27.61} & \textbf{0.855} & \textbf{0.176} &\\
\bottomrule[1.pt]
 \end{tabular}}
 \label{tab:joint}
\end{table}
 
\subsection{Frame2Model Registration}

In this section, current SOTA PE-based LATITUDE~\cite{latitude} and hash-encoding-based iNGP~\cite{liu2023baa} are compared with our method on varying levels of initial translation error in altitude on both real-world and simulation datasets. 

As shown in Table \ref{tab:poseestimate}, errors initially occur during the drone's ascent and descent, due to changes in the captured scene's resolution. A noticeable decline in LATITUDE's performance can be observed. This is attributable to the fact that LATITUDE's network architecture, derived from Mega-NeRF~\cite{meganerf}, cannot handle aliasing issues under different resolutions, leading to diminished performance. The previous centimeter-level registration achievement was due to the errors occurring while the drone was in level flight, with consistent imaging resolution for the scene.

For iNGP, its explicit encoding approach contributes to excellent overall image quality. However, some floaters affect the registration results. As can be observed, the method achieves better results with minor errors. Yet, the storage of model in this paper is 2.5GB, and it will exponentially increase with the scale of the scene.

Our method, termed iMNeRF, overall exhibits superior performance, especially in real-world data scenarios. This is mainly attributed to our refinement of the joint optimization of scene and pose in Mip-NeRF 360, which enhances the representation of the scene itself. 
Incorporating the TDLF in our method enhances robustness by suppressing high-frequency artifacts, particularly in managing perspectives from both higher and lower altitudes, outperforming other methods. Furthermore, the storage size of our model is only 103MB.

\begin{figure}[!t]
\vspace{1.5mm}
\centering
\includegraphics[width=0.48\textwidth]{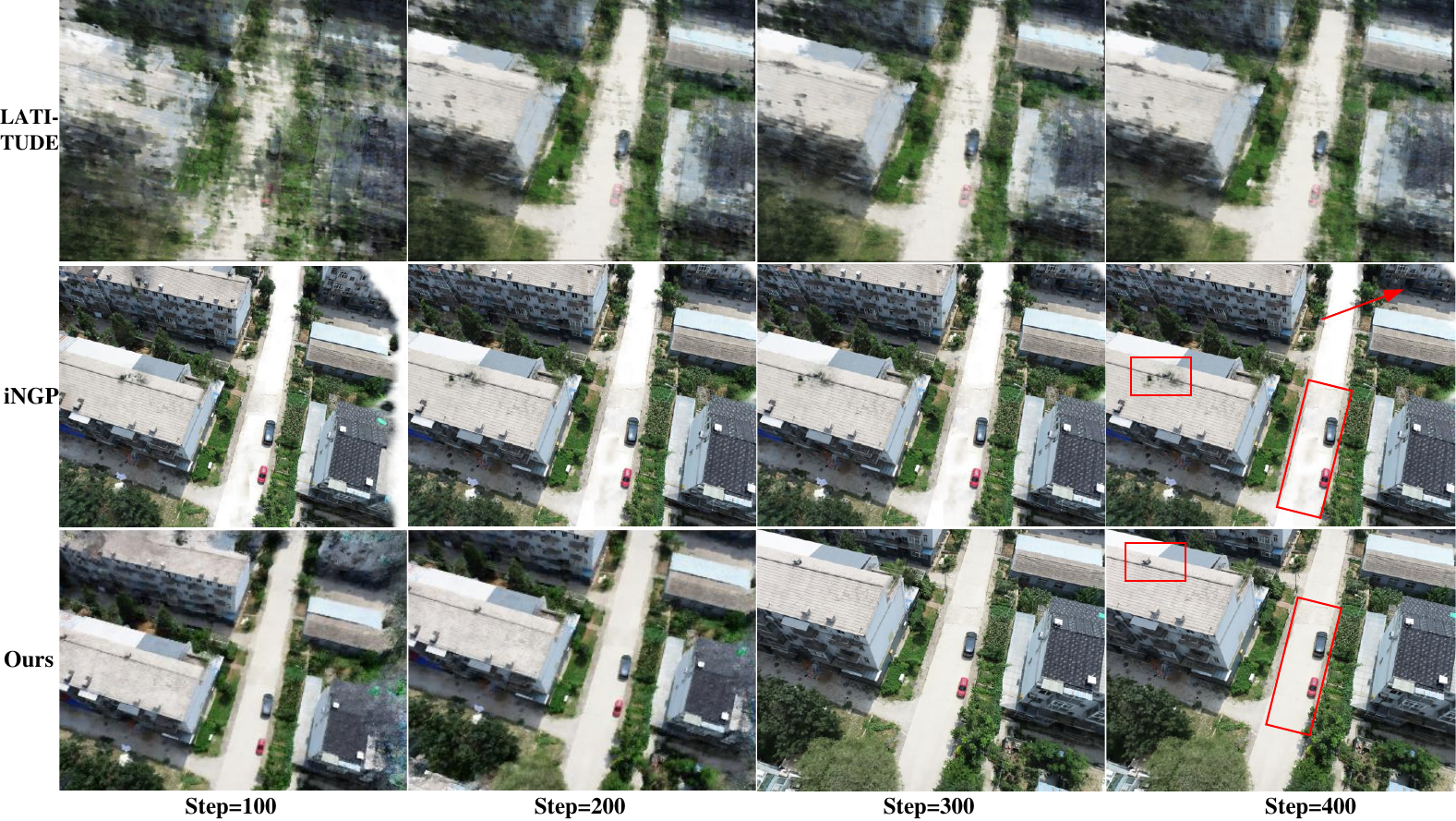}%
\caption{Frame2Model registration results: the LATITUDE\cite{latitude} experiences registration failure due to aliasing, resulting in confusion in the merged observation and rendering images. While the iNGP\cite{liu2023baa} shows notable accuracy improvements, noticeable floating artifacts in the highlighted area are observed, potentially inducing registration errors. Ours effectively avoids aliasing, ensuring superior rendering quality and registration accuracy.}
\label{registration}
\vspace{-6mm}
\end{figure}

Fig.\ref{registration} shows the registration process of these methods when the initial error is 10 meters. LATITUDE is incapable of addressing the aliasing issues resulting from resolution variations. Conversely, the presence of floaters in iNGP adversely impacts the quality of localization. Furthermore, solely relying on entire frequency domain information for pose optimization can inadvertently lead to local optima. In contrast, our approach excels at addressing challenges arising from image aliasing and complex signal interferences. As observed in the figure, initially, we suppressed a portion of the high-frequency information, leading to a reduction in image quality. However, utilizing both high and low-frequency information toward the end restored the image quality to a satisfactory level.

\begin{table*}[!t]
  \centering
  \caption{Translation and rotation errors of NeRF-based registration results}
  \small
  \renewcommand{\tabcolsep}{5pt}
 \resizebox{0.8\textwidth}{!}{
 \begin{tabular}{@{}clcccccccc@{}}
  \toprule[1pt]
  \multirow{2}{*}{\begin{tabular}[c]{@{}c@{}}Scene\end{tabular}}
   &\multirow{2}{*}{\begin{tabular}[c]{@{}c@{}}Method\end{tabular}}
   &\multicolumn{2}{c}{4}
   &\multicolumn{2}{c}{-4} 
   &\multicolumn{2}{c}{10}
   &\multicolumn{2}{c}{-10} \\
    \cmidrule{3-4}
    \cmidrule(l){5-6}
    \cmidrule(l){7-8}
    \cmidrule(l){9-10}
  & & Rotation($^\circ$) &Translation(m)
  & Rotation($^\circ$) &Translation(m)
  & Rotation($^\circ$) &Translation(m)
  & Rotation($^\circ$) &Translation(m) \\
\midrule
\multirow{3}{*}{\begin{tabular}[c]{@{}c@{}}Sim\end{tabular}} 
    &LATITUDE  & 0.9720 & 0.1200 & 7.9650 & 0.0210 & 12.602 & 0.3550 & 2.6820 & 0.4630\\
    &iNGP & \textbf{0} & \textbf{0.0040} & \textbf{0} & 0.0768 & 0.0470 & 4.2940 & 0.0720 & 8.3580 \\
    &Ours(iMNeRF)      & 0.0005 & 0.0291 & 0.0005 & \textbf{0.0118} & \textbf{0.0005} & \textbf{0.0189} & \textbf{0.0005} & \textbf{0.0214}\\
\midrule 
\multirow{3}{*}{\begin{tabular}[c]{@{}c@{}}Real\end{tabular}} 
   &LATITUDE  & 2.0710 & 2.4040 & 1.3260 & 0.7950 & 3.8430 & 6.9930 & 4.1260 & 5.2150\\
    &iNGP     & 0.0167 & 0.5270 & \textbf{0} & \textbf{0.0040} & 0.0860 & 5.3230 & 0.0160 & 2.9510 \\
    &Ours(iMNeRF)       & \textbf{0.0005} & \textbf{0.0111} & 0.0005 & 0.0058 & \textbf{0.0007} & \textbf{0.0063} & \textbf{0} & \textbf{0.0151}\\
\bottomrule[1.pt]
\end{tabular}}
 \label{tab:poseestimate}
\end{table*}

\subsection{NeRF Blending with Model2Model Registration}

\begin{figure*}[!t]
\centering
\includegraphics[width=0.65\textwidth]{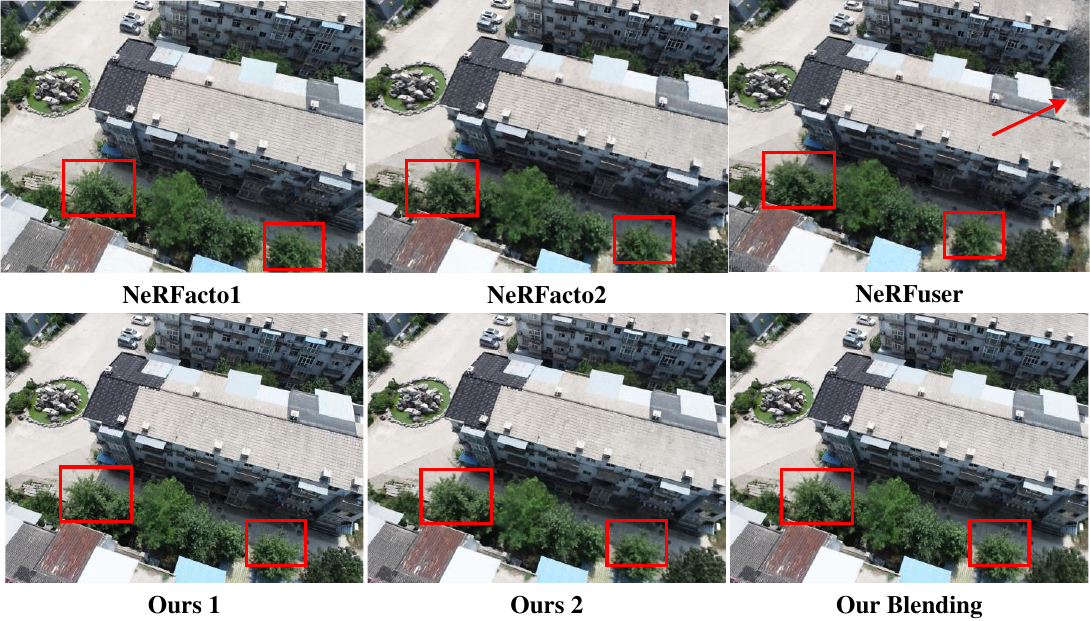}%
\caption{Blending Results in Real World. The first two columns represent the rendering results of distributed NeRFs trained by different methods. Due to data loss at the edges of the UAV trajectory, aliasing occurs in the rendering as indicated. When comparing renderings from the two blended NeRFs, our method significantly enhances areas where individual NeRF renderings perform poorly.}
\label{fusion_real}
\vspace{-4mm}
\end{figure*}

In a comprehensive evaluation of both real-world and simulation datasets, we compared with the blending results of NeRFuser. Specifically, our approach takes posed images in distinct NeRFs as input and produces NeRF rendering and blending results as output under the same setting as NeRFuser. Quantitative results, as detailed in Tables \ref{tab:fuser} and Table \ref{tab:ourfusion}, highlight a significant observation: for the same city-scale scenes, the individual NeRF reconstruction results by NeRFuser consistently underperform compared to ours.

\begin{table}[H]
  \centering
  \caption{NeRF Blending Results of NeRFuser~\cite{fang2023nerfuser}}
  \small
  \renewcommand{\tabcolsep}{5pt}
 \resizebox{0.28\textwidth}{!}{
 \begin{tabular}{@{}clccc@{}}
  \toprule[1pt]
   {\begin{tabular}[c]{@{}c@{}}Scene\end{tabular}}
   &{\begin{tabular}[c]{@{}c@{}}Methods\end{tabular}}
   &{PSNR$\uparrow$}
   &{SSIM$\uparrow$} 
   &{LPIPS$\downarrow$}  \\
\midrule
\multirow{3}{*}{\begin{tabular}[c]{@{}c@{}}Sim\end{tabular}} 
    &NeRFacto1    & 23.94 & 0.789 & 0.132 \\
    &NeRFacto2  & 23.77 & 0.792 & 0.134 \\
    &NeRFuser & 19.58 & 0.634 & 0.254 \\
\midrule 
\multirow{3}{*}{\begin{tabular}[c]{@{}c@{}}Real\end{tabular}} 
   &NeRFacto1    & 21.95 & 0.699 & 0.302 \\  
    &NeRFacto2  & 20.50 & 0.605 & 0.328 \\ 
    &NeRFuser  & 18.26 & 0.401 &0.399  \\
\bottomrule[1.pt]
\end{tabular}}
\label{tab:fuser}
\end{table}
\vspace{-0.1in}

\begin{table}[H]
  \centering
  \caption{NeRF Blending Results of Ours.}
  \small
  \renewcommand{\tabcolsep}{5pt}
 \resizebox{0.28\textwidth}{!}{
 \begin{tabular}{@{}clccc@{}}
  \toprule[1pt]
   {\begin{tabular}[c]{@{}c@{}}Scene\end{tabular}}
   &{\begin{tabular}[c]{@{}c@{}}Methods\end{tabular}}
    &{PSNR$\uparrow$}
   &{SSIM$\uparrow$} 
   &{LPIPS$\downarrow$}  \\
\midrule
\multirow{3}{*}{\begin{tabular}[c]{@{}c@{}}Sim\end{tabular}} 
    &Ours 1  &  26.43 & 0.825 & 0.306 \\
    &Ours 2  & 24.19 & 0.780 & 0.339 \\
    &Our Blending     & \textbf{25.58} & \textbf{0.812} &\textbf{0.317} \\ 
\midrule 
\multirow{3}{*}{\begin{tabular}[c]{@{}c@{}}Real\end{tabular}} 
   &Ours 1 & 27.61 & 0.855 & 0.176 \\
   &Ours 2 & 24.12 & 0.678 & 0.355 \\
   &Our Blending   & \textbf{26.52} & \textbf{0.820} & \textbf{0.195} \\   
\bottomrule[1.pt]
 \end{tabular}}
 \label{tab:ourfusion}
\end{table}
\vspace{0.7mm}

Utilizing Model2Model pose optimization for blending, our approach excels in integrating the most desirable features while adeptly addressing inherent limitations. In contrast, using NeRFuser for blending visibly reduces scene representation quality, indicating limitations in leveraging distributed agents' full potential. Thus, our method significantly outperforms in large-scale scene blending.

The Fig.\ref{fusion_real} show the qualitative result of NeRF blending. NeRFuser, which employs NeRFacto for training a distributed NeRF, exhibits registration inaccuracies. Thus images synthesized fail to align seamlessly within a unified coordinate system, leading to suboptimal blending. In contrast, our framework benefits from the excellent registration result after tri-stage optimization, achieving a better blending result. It can be observed that the our blended image effectively incorporates the strengths and compensates for the weaknesses of the two distributed images, resulting in improved performance.

\section{CONCLUSIONS}
In this paper, we propose a robust tri-stage pose optimization technique within a distributed NeRF system. This approach effectively addresses the challenges observed in previous NeRF registration stages. Through the strategic implementation of the bundle-adjusting Mip-NeRF 360, our system offers precise pose estimation for the images themselves. Further enhanced by the incorporation of the truncated dynamic low-pass filter, our iMNeRF achieves dependable and accurate Frame2Model registration. Building on the initial relative transformation after Frame2Model optimization, Model2Model registration is then executed. As a result, our system not only corrects occlusion artifacts during the NeRF blending process but also demonstrates significant performance enhancements in both real-world and simulation environments. 
Looking ahead, our future efforts will focus on air-ground collaboration to achieve more generalizable distributed scenes reconstruction.

\bibliographystyle{plain}  
\bibliography{ref} 
\end{document}